\documentclass{article}
\usepackage{spconf,amsmath,graphicx}

\usepackage{booktabs}
\usepackage{multirow}
\usepackage[table]{xcolor}
\definecolor{Highlight}{HTML}{39b54a}  
\definecolor{Cerulean}{HTML}{00A2E3}  

\usepackage{enumitem}
\usepackage{kotex} 
\usepackage{CJKutf8}

\title{Cross-Lingual Learning in Multilingual Scene Text Recognition}

\name{Jeonghun Baek, Yusuke Matsui, Kiyoharu Aizawa}
\address{The University of Tokyo}

\begin{document}
%
\maketitle

\begin{abstract}
In this paper, we investigate cross-lingual learning (CLL) for multilingual scene text recognition (STR).
CLL transfers knowledge from one language to another.
We aim to find the condition that exploits knowledge from high-resource languages for improving performance in low-resource languages.
To do so, we first examine if two general insights about CLL discussed in previous works are applied to multilingual STR: (1) Joint learning with high- and low-resource languages may reduce performance on low-resource languages, and (2) CLL works best between typologically similar languages.
Through extensive experiments, we show that two general insights may not be applied to multilingual STR. 
After that, we show that the crucial condition for CLL is the dataset size of high-resource languages regardless of the kind of high-resource languages.
Our code, data, and models are available at https://github.com/ku21fan/CLL-STR.
\end{abstract}


\section{Introduction}
Scene text recognition (STR) is a task of reading text written within word- or line-level scene images.
Most STR methods have been developed using monolingual English data~\cite{CRNN,ASTER,TRBA,STRfewerlabels,ABINet,SimAN,SVTR,PARSeq,TextAdaIN,ECCV2022mgp_str}.
Recently, multilingual STR, which reads text written in multiple languages, has been drawn to attention gradually~\cite{etter2023hybrid,MRN}.
Meanwhile, cross-lingual learning (CLL) in multilingual STR has not been investigated yet.

\begin{figure}[t]
\centering
    \includegraphics[width=\linewidth]{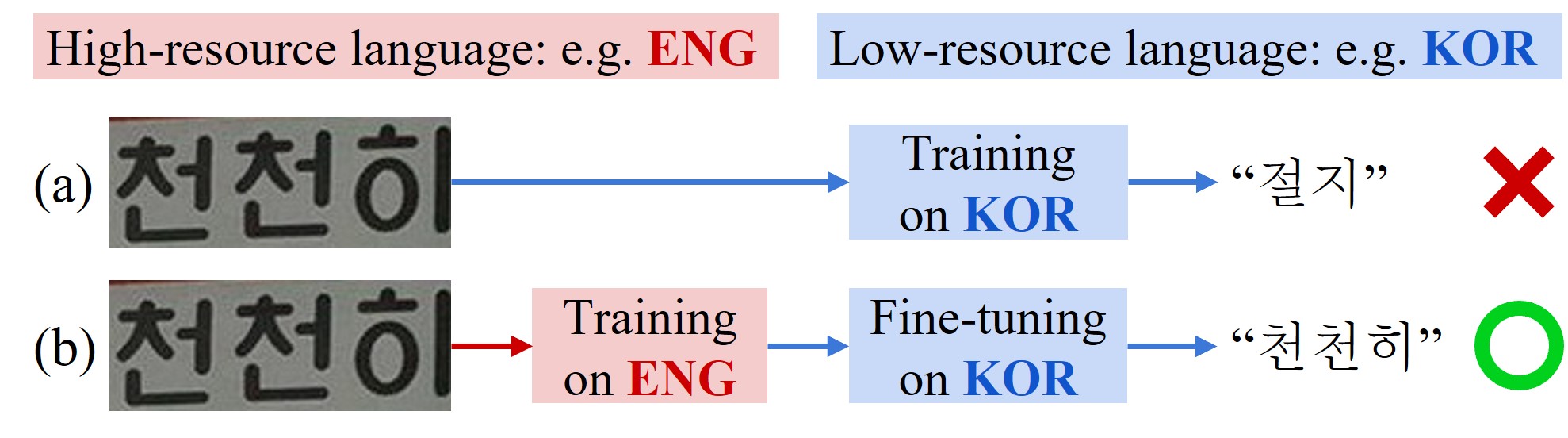}
    \vspace{-7mm}
    \caption{Scene text recognition in a low-resource language, Korean (KOR). 
    (a) Monolingual training on KOR results in incorrect predictions, whereas (b) CLL from a high-resource language English (ENG) to KOR leads to correct predictions.}
  \label{fig:CLL}
\end{figure}

CLL transfers knowledge from one language to another, or more broadly, from some multiple languages to other multiple languages. 
Often, the number of resources in each language is not even.
There are high-resource languages such as English and low-resource languages such as Korean.
Generally, the performance on a low-resource language is low, and improving the performance requires more resources.
In this case, instead of obtaining more resources, applying CLL from an existing high-resource language to the low-resource language can improve the performance, as shown in Figure~\ref{fig:CLL}.

In this paper, we investigate CLL in multilingual STR to exploit knowledge of high-resource languages for low-resource languages.
Particularly, we aim to find the condition where CLL works well.
To do so, we first verify two general insights about CLL discussed in previous papers~\cite{MRN,M-BERT}: (1) joint learning with high- and low-resource languages may reduce performance on low-resource languages~\cite{MRN}, and (2) CLL works best between typologically similar languages~\cite{M-BERT}.
These insights are not explicitly verified in multilingual STR yet.
By using a representative multilingual scene text dataset MLT19~\cite{MLT19}, we show that the two general insights may not be applied to multilingual STR.

After that, through extensive experiments, we show that the crucial factor is the number of samples in high-resource languages rather than the similarity between languages. 
In other words, we show that CLL works well when we have sufficiently large samples in high-resource languages regardless of the kind of high-resource languages.

\section{Cross-lingual learning in STR}\label{sec:cll}
\subsection{CLL methods for STR}
CLL is a kind of transfer learning that transfers knowledge from one language to another~\cite{CLLsurvey}.
The knowledge transfer is not strictly limited to one-to-one transfer but also some multiple languages to other multiple languages.
CLL has mainly been used for natural language processing (NLP) tasks with zero-shot cross-lingual model transfer: A pre-trained model on multiple languages is fine-tuned using task-specific data from one language and evaluated in other languages.
For example, in the multilingual BERT (M-BERT) paper~\cite{M-BERT}, the authors evaluate CLL performance via zero-shot cross-lingual model transfer for named entity recognition (NER)~\cite{tjong-kim-sang-de-meulder-2003-introduction} and part of speech (POS) tagging task~\cite{nivre-etal-2016-universal}.

Unlike NER and POS tasks, using the zero-shot cross-lingual model transfer for STR is nearly impossible.
This is because STR aims to read the characters in the image, whereas NER and POS tasks aim to predict the role of each word in the sentence.
Reading characters never seen during training is nearly impossible.
Therefore, we need other ways to evaluate CLL performance for STR rather than zero-shot cross-lingual transfer.

In this paper, we investigate the effect of CLL via joint and cascade learning.
They are simple and basic methods for CLL.
Joint learning is to train a model using multiple datasets from multiple languages simultaneously. 
Cascade learning is to train a model in one language and then fine-tune the model in another language~\cite{CLLsurvey}.

\begin{figure}[t]
\centering
    \includegraphics[width=\linewidth]{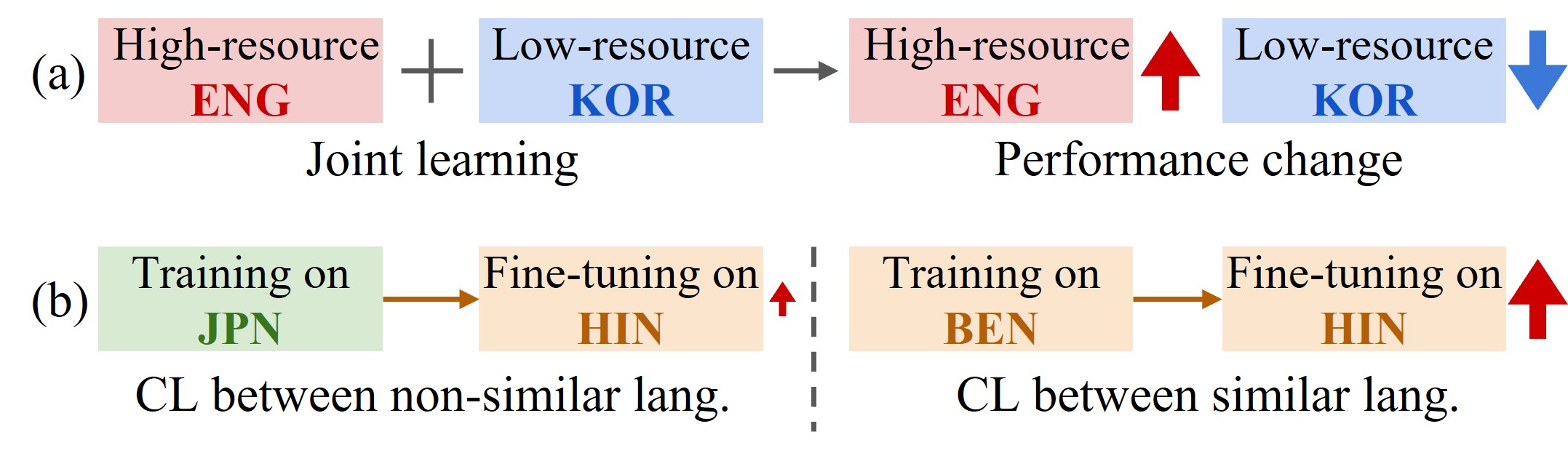}
    \vspace{-9mm}
    \caption{We investigate two general insights about CLL related to (a) joint learning and (b) cascade learning (CL).}
    \label{fig:insights}
    \vspace{-2mm}
\end{figure}

\subsection{Two general insights about CLL}
Figure~\ref{fig:insights} illustrates two general insights related to CLL for STR: One is from the recent multilingual STR work MRN~\cite{MRN}, and another is from M-BERT~\cite{M-BERT}.
According to MRN~\cite{MRN}, joint learning on high- and low-resource languages may make the model biased toward high-resource languages, resulting in poor performance on low-resource languages.
Since the model has a limited capacity, the model can be biased to the data from high-resource languages.
However, this insight is not explicitly verified yet. 
The authors of MRN did not show results related to it. 
In this work, we explicitly verify this insight.

According to M-BERT~\cite{M-BERT}, CLL works best when the linguistic typology (e.g., the order of subject, verb, and object in a sentence) of source and target languages are similar.
Regardless of linguistic typology, the essential part of this insight is the \textit{similarity} between languages.
For the STR task, similarity in appearance (e.g., shape or stroke) may be more important than linguistic typology since an STR sample is usually a word-level image; thus, linguistic typology's influence may decrease compared to other NLP tasks.
We investigate CLL performance with cascade learning between similar languages in linguistic typology and appearance.

\begin{table}[t] 
\tabcolsep=0.12cm
    \centering
    \begin{tabular}{@{}lrrrrrrr@{}}
        \toprule
        Count type & CHI & BEN & HIN & JPN & ARA & KOR & LAT \\
        \midrule
        Training & 1.9K & 2.6K & 2.7K & 3.1K & 3.5K & 4.0K & 28K \\
        Validation & 204 & 341 & 320 & 337 & 408 & 455 & 3.4K \\
        Evaluation & 248 & 311 & 349 & 366 & 421 & 525 & 3.3K \\
        Charset & 1.6K & 608 & 696 & 1.3K & 141 & 878 & 85 \\
        \midrule
        SynthMLT & 33K & 165K & 54K & 37K & 202K & 147K & 211K \\
        \bottomrule
    \end{tabular}
    \vspace{-2mm}
    \caption{Statistics of the multilingual STR datasets.
    The charset is counted from the training set.
    }
    \label{tab:data}
    \vspace{-2mm}
\end{table}     

\begin{table*}[t] 
  \tabcolsep=0.13cm
    \begin{center}
        \begin{tabular}{@{}l|ccccccc|c|ccccccc|c@{}}
            \toprule
            & \multicolumn{8}{c|}{Model: CRNN (TPAMI’17)~\cite{CRNN}} & \multicolumn{8}{c}{Model: SVTR (IJCAI’22)~\cite{SVTR}} \\ 
            Training data & CHI & BEN & HIN & JPN & ARA & KOR & \multicolumn{1}{c}{LAT} & Avg. & CHI & BEN & HIN & JPN & ARA & KOR & \multicolumn{1}{c}{LAT} & Avg. \\
            \midrule
            Each lang. (\textit{Mono.}) & 3.2 & 33.4 & 24.2 & 14.3 & 27.1 & 4.3 & \textbf{85.7} & 27.4 & 2.1 & 19.1 & 22.8 & 10.9 & 23.6 & 5.7 & \textbf{85.0} & 24.2 \\
            All lang. (\textit{All}, 46K) & \textbf{41.3} & \textbf{69.3} & \textbf{66.9} & \textbf{44.8} & \textbf{54.7} & \textbf{61.4} & 85.4 & \textbf{60.5} & \textbf{35.2} & \textbf{67.8} & \textbf{68.5} & \textbf{42.9} & \textbf{53.7} & \textbf{53.0} & 84.7 & \textbf{58.0} \\
            \textit{All but LAT} (18K) & 8.3 & 32.6 & 31.6 & 17.8 & 36.9 & 12.2 & $-$ & $-$ & 4.1 & 19.5 & 22.9 & 12.5 & 26.5 & 10.1 & $-$ & $-$ \\
            \bottomrule
        \end{tabular}
    \vspace{-2mm}
    \caption{Results of joint learning.
    The higher the better.
    The number in parenthesis denotes the number of samples.
    }
    \label{tab:joint}
    \end{center}
    \vspace{-3mm}
\end{table*}

\section{Multilingual scene text datasets}
We use the representative multilingual scene text dataset, MLT19~\cite{MLT19}.
MLT19 consists of scene images from 10 languages: Arabic, Bengali, Chinese, Devanagari, English, French, German, Italian, Japanese, and Korean.
MLT19 was released at the ICDAR2019 competition, and from the competition, researchers used 8 categories for evaluation: Arabic (ARA), Latin (LAT), Chinese (CHI), Japanese (JPN), Korean (KOR), Bengali (BEN), Hindi (HIN), and Symbols.
LAT consists of English, French, German, and Italian. 

In our experiments, to strictly verify CLL, we would like to minimize the character overlap between categories (languages).
In each category, we exclude the samples containing characters from other categories, symbols, and digits.
As a result, we do not use the Symbols category: we use 7 categories for experiments.
The character overlap only exists between CHI and JPN: CHI and JPN share 602 Chinese characters.

Table~\ref{tab:data} shows the statistics of MLT19. 
The order of languages is sorted by the number of the training set.
Note that the character set (charset) differs in each language.
In subsection~\ref{crucial}, we also use the well-known synthetic data, synthetic MLT (SynthMLT)~\cite{SynthMLT}, which was released along with MLT19. 
To strictly verify CLL, SynthMLT samples containing characters that do not exist in MLT19 are excluded.

\begin{table*}[t] 
  \tabcolsep=0.13cm
    \begin{center}
        \begin{tabular}{@{}l|ccccccc|ccccccc@{}}
            \toprule
            & \multicolumn{7}{c|}{Model: CRNN (TPAMI’17)~\cite{CRNN}} & \multicolumn{7}{c}{Model: SVTR (IJCAI’22)~\cite{SVTR}} \\
            Pre-train data & CHI & BEN & HIN & JPN & ARA & KOR & LAT & CHI & BEN & HIN & JPN & ARA & KOR & LAT \\
            \midrule
            None (\textit{Mono.}) & 3.2 & 33.4 & 24.2 & 14.3 & 27.1 & 4.3 & 85.7 & 2.1 & 19.1 & 22.8 & 10.9 & 19.1 & 5.7 & \textbf{85.0} \\
            \textit{All} (46K) & \textbf{44.3} & \textbf{71.8} & \textbf{70.2} & \textbf{47.7} & \textbf{55.0} & \textbf{63.0} & 85.6 & \textbf{36.3} & \textbf{73.4} & \textbf{73.8} & \textbf{43.9} & \textbf{55.3} & 56.4 & 84.9 \\ 
            \textit{All but LAT} (18K) & 9.2 & 34.5 & 35.0 & 19.7 & 37.0 & 13.3 & 77.6 & 5.3 & 23.7 & 26.0 & 14.1 & 27.9 & 11.1 & 81.0 \\ 
            \midrule
            CHI (1.9K) & $-$ & 19.8 & 21.5 & 11.9 & 20.5 & 6.4 & \textbf{86.5} & $-$ & 17.2 & 18.7 & 8.5 & 19.8 & 5.2 & 80.9  \\
            BEN (2.6K) & 3.6 & $-$ & 32.2 & 12.2 & 20.4 & 6.4 & 80.0 & 2.3 & $-$ & 22.5 & 11.1 & 23.3 & 5.9 & 80.3  \\
            HIN (2.7K) & 2.1 & 25.5 & $-$ & 11.4 & 24.0 & 8.8 & 85.9  & 3.2 & 24.8 & $-$ & 10.8 & 23.4 & 7.8 & 80.1  \\
            JPN (3.1K) & 1.5 & 15.0 & 26.1 & $-$ & 14.7 & 5.9 & 84.4 & 1.7 & 19.5 & 20.7 & $-$ & 20.5 & 7.1 & 80.2  \\
            ARA (3.5K) & 2.9 & 26.4 & 23.9 & 13.8 & $-$ & 4.6 & 86.0 & 1.7 & 16.6 & 18.3 & 9.8 & $-$ & 4.5 & 80.8  \\
            KOR (4.0K) & 1.2 & 19.6 & 20.7 & 11.9 & 21.2 & $-$ & 85.9 & 3.8 & 18.3 & 20.6 & 11.2 & 18.4 & $-$ & 80.4 \\
            LAT (28K) & 18.2 & 52.9 & 69.2 & 32.1 & 28.4 & 44.6 & $-$ & 28.6 & 67.3 & 69.9 & 43.4 & 53.2 & \textbf{57.1} & $-$ \\
            \arrayrulecolor{lightgray}\hline\arrayrulecolor{black}
            2K LAT & 3.4 & 25.2 & 27.3 & 13.1 & 17.5 & 11.7 & $-$ & 3.2 & 15.8 & 18.8 & 10.3 & 19.5 & 5.2 & $-$ \\
            \bottomrule
        \end{tabular}
    \vspace{-2mm}
    \caption{
    Results of cascade learning: Pre-train a model and then fine-tune the model on each language. 
    }
    \label{tab:FT}
    \end{center}
    \vspace{-3mm}
\end{table*}

\begin{table*}[t] 
  \tabcolsep=0.12cm
    \begin{center}
        \begin{tabular}{@{}l|ccccccc|c|ccccccc|c@{}}
            \toprule
            & \multicolumn{8}{c|}{Model: CRNN (TPAMI’17)~\cite{CRNN}} & \multicolumn{8}{c}{Model: SVTR (IJCAI’22)~\cite{SVTR}} \\ 
            Training data & CHI & BEN & HIN & JPN & ARA & KOR & \multicolumn{1}{c}{LAT} & Avg. & CHI & BEN & HIN & JPN & ARA & KOR & \multicolumn{1}{c}{LAT} & Avg. \\
            \midrule
            \textit{Base} (20K) & 9.7 & 35.1 & 31.8 & 19.5 & 40.2 & 15.9 & 20.9 & 24.7 & 6.0 & 23.0 & 25.0 & 14.5 & 29.3 & 11.8 & 20.2 & 18.5 \\
            \arrayrulecolor{lightgray}\hline\arrayrulecolor{black}
            +~~~~2K S-CHI (22K) & 20.1 & 51.8 & 49.7 & 28.0 & 42.8 & 35.5 & 42.2 & 38.6 & 6.5 & 19.3 & 25.6 & 14.3 & 27.1 & 11.2 & 18.9 & 17.6 \\
            +~~33K S-CHI (53K) & \textbf{54.8} & 69.0 & 68.6 & 39.2 & 59.7 & 52.3 & 59.4 & 57.6 & \textbf{56.6} & 62.3 & 63.3 & 41.5 & 50.1 & 49.2 & 57.2 & 54.3 \\
            \arrayrulecolor{lightgray}\hline\arrayrulecolor{black}
            +~~~~2K S-BEN (22K) & 10.3 & 40.9 & 37.6 & 19.0 & 40.2 & 15.9 & 23.0 & 26.7 & 6.5 & 23.4 & 27.6 & 14.3 & 27.4 & 13.8 & 20.4 & 19.1 \\
            +165K S-BEN (185K) & 32.7 & \textbf{81.3} & 74.2 & 40.1 & 65.1 & 55.1 & 63.2 & 58.8 & 44.0 & \textbf{84.8} & \textbf{80.8} & 51.0 & 70.6 & 63.2 & 68.7 & 66.2 \\
            \arrayrulecolor{lightgray}\hline\arrayrulecolor{black}
            +~~~~2K S-HIN (22K) & 17.7 & 55.1 & 52.3 & 28.7 & 41.7 & 36.0 & 42.5 & 39.2 & 5.8 & 22.3 & 26.3 & 13.9 & 28.1 & 11.1 & 19.2 & 18.1 \\
            +~~54K S-HIN (74K) & 36.4 & 74.2 & \textbf{78.7} & 41.4 & 57.7 & 57.2 & 61.2 & 58.1 & 29.0 & 64.6 & 70.0 & 37.4 & 52.7 & 44.6 & 55.3 & 50.5 \\
            \arrayrulecolor{lightgray}\hline\arrayrulecolor{black}
            +~~~~2K S-JPN (22K) & 19.4 & 55.9 & 53.9 & 30.5 & 41.8 & 39.2 & 43.5 & 40.6 & 6.0 & 20.9 & 25.2 & 14.6 & 27.3 & 12.7 & 19.0 & 18.0 \\
            +~~37K S-JPN (57K) & 39.2 & 70.3 & 71.5 & \textbf{54.8} & 59.3 & 59.8 & 63.7 & 59.8 & 38.6 & 66.9 & 70.2 & \textbf{56.7} & 55.3 & 53.9 & 62.7 & 57.7 \\
            \arrayrulecolor{lightgray}\hline\arrayrulecolor{black}
            +~~~~2K S-ARA (22K) & 13.5 & 46.4 & 43.9 & 26.4 & 41.9 & 31.6 & 39.4 & 34.7 & 6.4 & 19.9 & 25.8 & 14.3 & 28.8 & 13.1 & 19.4 & 18.2 \\
            +202K S-ARA (222K) & 32.4 & 71.1 & 72.1 & 37.2 & \textbf{82.2} & 54.0 & 58.7 & 58.2 & 33.7 & 69.2 & 70.8 & 40.8 & \textbf{82.7} & 50.0 & 57.4 & 57.8 \\
            \arrayrulecolor{lightgray}\hline\arrayrulecolor{black}
            +~~~~2K S-KOR (22K) & 10.2 & 37.4 & 35.0 & 19.6 & 38.8 & 19.4 & 20.7 & 25.9 & 6.4 & 23.2 & 25.8 & 14.0 & 28.2 & 14.9 & 20.9 & 19.1  \\
            +147K S-KOR (167K) & 40.9 & 75.2 & 73.6 & 45.7 & 65.2 & \textbf{80.4} & \textbf{64.3} & \textbf{63.6} & 50.7 & 81.0 & 78.4 & 53.5 & 73.6 & \textbf{82.5} & \textbf{69.6} & \textbf{69.9} \\
            \bottomrule
        \end{tabular}
    \vspace{-4mm}
    \caption{Joint learning with SynthMLT~\cite{SynthMLT}.
    \textit{Base} is 2K LAT and six other languages.
    The prefix ``S-'' denotes SynthMLT data.
    }
    \label{tab:joint-syn}
    \end{center}
    \vspace{-3mm}
\end{table*}

\begin{table*}[t] 
  \tabcolsep=0.13cm
    \begin{center}
        \begin{tabular}{@{}l|ccccccc|ccccccc@{}}
            \toprule
            & \multicolumn{7}{c|}{Model: CRNN (TPAMI’17)~\cite{CRNN}} & \multicolumn{7}{c}{Model: SVTR (IJCAI’22)~\cite{SVTR}} \\
            Pre-train data & CHI & BEN & HIN & JPN & ARA & KOR & LAT & CHI & BEN & HIN & JPN & ARA & KOR & LAT \\
            \midrule
            ~~~~2K S-CHI & $-$ & 20.7 & 21.9 & 14.2 & 22.5 & 7.7 & \textbf{86.4} &  $-$ & 12.0 & 16.0 & 6.5 & 20.3 & 3.6 & 80.4 \\
            ~~33K S-CHI  & $-$ & 42.1 & 55.6 & 26.1 & \textbf{36.3} & 36.8 & 80.9 & $-$ & 51.8 & 53.9 & 31.5 & 44.8 & 43.0 & 83.4 \\
            \arrayrulecolor{lightgray}\hline\arrayrulecolor{black}
            ~~~~2K S-BEN & 2.6 & $-$ & 22.8 & 9.6 & 16.9 & 3.0 & 78.0 & 2.2 & $-$ & 18.3 & 9.0 & 20.9 & 4.5 & 80.8 \\
            165K S-BEN  & 13.0 & $-$ & \textbf{72.7} & 20.8 & 33.9 & 40.4 & 84.6 & 41.2 & $-$ & \textbf{84.8} & 50.6 & 72.2 & \textbf{69.2} & 88.3 \\
            \arrayrulecolor{lightgray}\hline\arrayrulecolor{black}
            ~~~~2K S-HIN & 3.0 & 19.8 & $-$ & 12.7 & 20.3 & 6.9 & 86.2 & 2.1 & 14.0 & $-$ & 9.0 & 20.1 & 4.5 & 80.8 \\
            ~~54K S-HIN  & 7.4 & 66.6 & $-$ & 15.4 & 32.1 & 31.7 & 82.3 & 17.2 & 64.5 & $-$ & 29.6 & 48.6 & 39.4 & 84.0 \\
            \arrayrulecolor{lightgray}\hline\arrayrulecolor{black}
            ~~~~2K S-JPN & 2.9 & 20.0 & 23.9 & $-$ & 19.1 & 7.8 & 86.1 & 1.5 & 15.9 & 15.0 & $-$ & 18.4 & 4.4 & 80.9 \\
            ~~37K S-JPN  & \textbf{25.3} & 44.1 & 47.5 & $-$ & 36.2 & \textbf{41.4} & 82.7 & 33.5 & 68.6 & 70.1 & $-$ & 58.6 & 55.2 & 85.4 \\
            \arrayrulecolor{lightgray}\hline\arrayrulecolor{black}
            ~~~~2K S-ARA & 2.4 & 18.1 & 22.0 & 10.4 & $-$ & 5.8 & 85.2 & 1.4 & 14.6 & 13.0 & 8.2 & $-$ & 3.6 & 79.9 \\
            202K S-ARA & 11.6 & 57.5 & 59.6 & 25.3 & $-$ & 33.4 & 85.1 & 19.4 & 73.5 & 68.4 & 34.4 & $-$ & 56.8 & 87.6 \\
            \arrayrulecolor{lightgray}\hline\arrayrulecolor{black}
            ~~~~2K S-KOR & 2.2 & 24.6 & 22.7 & 10.7 & 23.3 & $-$ & 75.8 & 1.6 & 14.7 & 16.8 & 8.5 & 18.3 & $-$ & 80.6 \\
            147K S-KOR & 19.3 & \textbf{71.9} & 64.1 & \textbf{36.9} & 35.0 & $-$ & 84.8 & \textbf{52.8} & \textbf{82.4} & \textbf{84.8} & \textbf{54.9} & \textbf{79.8} & $-$ & \textbf{88.5} \\
            \bottomrule
        \end{tabular}
    \vspace{-2mm}
    \caption{Cascade learning with SynthMLT~\cite{SynthMLT}.
    The prefix ``S-'' denotes SynthMLT data.
    }
    \label{tab:FT-syn}
    \end{center}
    \vspace{-5mm}
\end{table*}

\section{Experiment and analysis}
\subsection{Implementation details}
Following a previous multilingual STR work MRN~\cite{MRN}, we use traditional CNN-based model CRNN~\cite{CRNN} and ViT~\cite{ViT}-based model SVTR~\cite{SVTR}. 
According to a cross-lingual language model (XLM-R~\cite{XLMR}) paper, the CLL performance depends on the model capacity: the larger, the better.
Different charsets between languages result in differences in model capacity, confusing the verification of CLL. 
To keep the model capacity across all languages, we use the union of charsets from all languages. 
The size of the united charset is 4,676. 
We use it for all experiments.
The model is trained with 100K iterations for joint learning/pre-training and 10K iterations for fine-tuning. 
The batch size is 128.
Following ICDAR competitions~\cite{ArT,ReCTS,MLT19}, we use the normalized edit distance (NED), which normalizes the edit distance to the range between 0 and 1, as the evaluation metric. 
Without specific mention, we use the average NED (\%) for comparison. 

\subsection{Verifying two general insights about CLL}\label{verify}
We verify two general insights as mentioned in \S\ref{sec:cll}.
Table~\ref{tab:joint} shows the results of joint learning.
Comparing the monolingual results (\textit{Mono.}) and joint learning results (\textit{All}), joint learning improves performance: 27.4\% vs. 60.5\% for CRNN and 24.2\% vs. 58.0\% for SVTR.
Only LAT performance slightly decreases by -0.3\%.
Note that LAT is the most high-resource language in MLT19.
These results show that the first general insight is not applied to STR.
\textit{Simple joint learning, even without specific modules for CLL, drastically improves performance in low-resource languages.}

Regarding the second general insight, Table~\ref{tab:FT} shows the results of cascade learning.
Similar to joint learning, pre-training on all languages (\textit{All}) improves performance in most languages except LAT.
We also investigate one-to-one cascade learning to verify the second general insight.
First of all, pre-training on LAT improves performance in all other languages.
This may be because the number of LAT samples is larger than others. 
We investigate the effect of the dataset size in \S\ref{crucial}.
Following M-BERT~\cite{M-BERT}, we group languages from the perspective of linguistic typology (order of subject, object, and verb)~\cite{wals}: (1) SVO: CHI, ARA, and LAT, and (2) SOV: BEN, HIN, JPN, and KOR.
In the case of CHI $\rightarrow$ ARA, the performance decreases for CRNN, compared to monolingual: 27.1\% $\rightarrow$ 20.5\%. 
For JPN/KOR/HIN $\rightarrow$ BEN, the performance also decreases for CRNN: 33.4\% $\rightarrow$ 15.0\%, 19.6\%, and 25.5\% respectively.
Although in the case of BEN $\rightarrow$ HIN, the performance improves for CRNN (24.2\% $\rightarrow$ 32.2\%), in most other cases, the performance slightly improves or decreases.

From the perspective of appearance (e.g., shape or stroke), there may be two groups: (1) CHI, JPN, and KOR, and (2) BEN, HIN, and ARA.
For the former group, most characters's width and height are the same.
They are not extremely thin characters such as ``1'' or ``l''.
For the latter group, characters are usually linked with a horizontal line (Shirorekha in HIN).
Cascade learning between CHI, JPN, and KOR slightly improves or decreases performance, but there is no significant difference.
In the case of ARA $\rightarrow$ BEN, the performance even decreases: 33.4\% $\rightarrow$ 26.4\% for CRNN and 19.1\% $\rightarrow$ 16.6\% for SVTR.
These results indicate the second general insight is not applied to STR: Similarity in linguistic typology or appearance may not be a crucial factor for CLL in STR.

\subsection{Crucial factor for better CLL performance}\label{crucial}
In this subsection, we aim to find the crucial factor for CLL in STR.
The second general insight indicates the crucial factor is the similarity between languages.
However, we showed it is not the case for STR.
In the former subsection, we show that cascade learning from LAT to others is significantly effective.
From the Tables~\ref{tab:joint} and \ref{tab:FT}, excluding LAT from \textit{All} (\textit{All but LAT}) decreases performance drastically.
These results indicate LAT is crucial for CLL.
Then, two questions arise: (1) Is LAT language crucial? and (2) Is the dataset size crucial?

To answer the question (1), we conduct experiments using only 2K LAT.
From Table~\ref{tab:FT}, pre-training on 2K LAT has much lower performance than pre-training on 28K LAT.
For joint learning, we conduct an experiment using 2K LAT and six other languages (\textit{Base} in Table~\ref{tab:joint-syn}).
Comparing \textit{Base} and \textit{All but LAT} in Table~\ref{tab:joint}, adding 2K LAT slightly improves performance but does not help much.
The performance of \textit{Base} is lower than monolingual results: 24.7\% vs. 27.4\% for CRNN and 18.5\% vs. 24.2\% for SVTR.
These results indicate the LAT language is not the crucial factor.

To answer the question (2), we conduct additional experiments with SynthMLT~\cite{SynthMLT}.
Tables~\ref{tab:joint-syn} and \ref{tab:FT-syn} show the results of joint learning and cascade learning, respectively: the prefix ``S-'' denotes SynthMLT data.
From Table~\ref{tab:joint-syn}, for each language, adding 2K SynthMLT samples to \textit{Base} does not improve performance much.
Although the performance of CRNN increases by about 15\% with 2K S-CHI, 2K S-HIN, and 2K S-JPN, there is still room for improvement.
Meanwhile, adding all SynthMLT samples for each language to \textit{Base} drastically improves performance, over +30\%. 
The improvement is maximum with 147K S-KOR, by +38.9\% for CRNN and +51.4\% for SVTR.
Similar tendencies also occur in cascade learning, as shown in Table~\ref{tab:FT-syn}. 
Comparing pre-training on 2K SynthMLT samples and on all SynthMLT samples, the former slightly increases or decreases performance, whereas the latter drastically improves performance. 
These results indicate that \textit{the crucial factor is the dataset size of high-resource languages, and CLL in STR is less dependent on the kind of high-resource languages.}
We assume this is because the essential knowledge of STR is distinguishing text in the image and can be learned from any language.

\section{Conclusion}
We have investigated cross-lingual learning (CLL) in STR and showed that two general insights may not be applied to the STR task.
After that, we have shown that the crucial factor for CLL can be the dataset size of high-resource languages regardless of the kind of high-resource languages.
This work is a stepping-stone for multilingual STR. 
We hope this work helps get more insights into multilingual STR.

\bibliographystyle{IEEEbib}
\bibliography{strings,refs}

\begin{thebibliography}{10}

\bibitem{CRNN}
Baoguang Shi, Xiang Bai, and Cong Yao,
\newblock ``An end-to-end trainable neural network for image-based sequence
  recognition and its application to scene text recognition,''
\newblock {\em TPAMI}, 2016.

\bibitem{ASTER}
Baoguang Shi, Mingkun Yang, Xinggang Wang, Pengyuan Lyu, Cong Yao, and Xiang
  Bai,
\newblock ``Aster: An attentional scene text recognizer with flexible
  rectification,''
\newblock {\em TPAMI}, 2018.

\bibitem{TRBA}
Jeonghun Baek, Geewook Kim, Junyeop Lee, Sungrae Park, Dongyoon Han, Sangdoo
  Yun, Seong~Joon Oh, and Hwalsuk Lee,
\newblock ``What is wrong with scene text recognition model comparisons?
  dataset and model analysis,''
\newblock in {\em ICCV}, 2019.

\bibitem{STRfewerlabels}
Jeonghun Baek, Yusuke Matsui, and Kiyoharu Aizawa,
\newblock ``What if we only use real datasets for scene text recognition?
  toward scene text recognition with fewer labels,''
\newblock in {\em CVPR}, 2021.

\bibitem{ABINet}
Shancheng Fang, Hongtao Xie, Yuxin Wang, Zhendong Mao, and Yongdong Zhang,
\newblock ``Read like humans: Autonomous, bidirectional and iterative language
  modeling for scene text recognition,''
\newblock in {\em CVPR}, 2021.

\bibitem{SimAN}
Canjie Luo, Lianwen Jin, and Jingdong Chen,
\newblock ``Siman: Exploring self-supervised representation learning of scene
  text via similarity-aware normalization,''
\newblock in {\em CVPR}, 2022.

\bibitem{SVTR}
Yongkun Du, Zhineng Chen, Caiyan Jia, Xiaoting Yin, Tianlun Zheng, Chenxia Li,
  Yuning Du, and Yu-Gang Jiang,
\newblock ``Svtr: Scene text recognition with a single visual model,''
\newblock in {\em IJCAI}, 2022.

\bibitem{PARSeq}
Darwin Bautista and Rowel Atienza,
\newblock ``Scene text recognition with permuted autoregressive sequence
  models,''
\newblock in {\em ECCV}, 2022.

\bibitem{TextAdaIN}
Oren Nuriel, Sharon Fogel, and Ron Litman,
\newblock ``Textadain: Paying attention to shortcut learning in text
  recognizers,''
\newblock in {\em ECCV}, 2022.

\bibitem{ECCV2022mgp_str}
Peng Wang, Cheng Da, and Cong Yao,
\newblock ``Multi-granularity prediction for scene text recognition,''
\newblock in {\em ECCV}, 2022.

\bibitem{etter2023hybrid}
David Etter, Cameron Carpenter, and Nolan King,
\newblock ``A hybrid model for multilingual ocr,''
\newblock in {\em ICDAR}, 2023.

\bibitem{MRN}
Tianlun Zheng, Zhineng Chen, BingChen Huang, Wei Zhang, and Yu-Gang Jiang,
\newblock ``Mrn: Multiplexed routing network for incremental multilingual text
  recognition,''
\newblock in {\em ICCV}, 2023.

\bibitem{M-BERT}
Telmo Pires, Eva Schlinger, and Dan Garrette,
\newblock ``How multilingual is multilingual {BERT}?,''
\newblock in {\em ACL}, 2019.

\bibitem{MLT19}
Nibal Nayef, Yash Patel, Michal Busta, Pinaki~Nath Chowdhury, Dimosthenis
  Karatzas, Wafa Khlif, Jiri Matas, Umapada Pal, Jean-Christophe Burie,
  Cheng-lin Liu, et~al.,
\newblock ``Icdar2019 robust reading challenge on multi-lingual scene text
  detection and recognition—rrc-mlt-2019,''
\newblock in {\em ICDAR}, 2019.

\bibitem{CLLsurvey}
Mat{\'u}{\v{s}} Pikuliak, Mari{\'a}n {\v{S}}imko, and M{\'a}ria Bielikov{\'a},
\newblock ``Cross-lingual learning for text processing: A survey,''
\newblock {\em ESWA}, 2021.

\bibitem{tjong-kim-sang-de-meulder-2003-introduction}
Erik~F. Tjong Kim~Sang and Fien De~Meulder,
\newblock ``Introduction to the {C}o{NLL}-2003 shared task:
  Language-independent named entity recognition,''
\newblock in {\em CoNLL}, 2003.

\bibitem{nivre-etal-2016-universal}
Joakim Nivre, Marie-Catherine de~Marneffe, Filip Ginter, Yoav Goldberg, Jan
  Haji{\v{c}}, Christopher~D. Manning, Ryan McDonald, Slav Petrov, Sampo
  Pyysalo, Natalia Silveira, Reut Tsarfaty, and Daniel Zeman,
\newblock ``{U}niversal {D}ependencies v1: A multilingual treebank
  collection,''
\newblock in {\em LREC}, 2016.

\bibitem{SynthMLT}
Michal Bu{\v{s}}ta, Yash Patel, and Jiri Matas,
\newblock ``E2e-mlt-an unconstrained end-to-end method for multi-language scene
  text,''
\newblock in {\em ACCV}, 2018.

\bibitem{ViT}
Alexey Dosovitskiy, Lucas Beyer, Alexander Kolesnikov, Dirk Weissenborn,
  Xiaohua Zhai, Thomas Unterthiner, Mostafa Dehghani, Matthias Minderer, Georg
  Heigold, Sylvain Gelly, Jakob Uszkoreit, and Neil Houlsby,
\newblock ``An image is worth 16x16 words: Transformers for image recognition
  at scale,''
\newblock in {\em ICLR}, 2021.

\bibitem{XLMR}
Alexis Conneau, Kartikay Khandelwal, Naman Goyal, Vishrav Chaudhary, Guillaume
  Wenzek, Francisco Guzm{\'a}n, Edouard Grave, Myle Ott, Luke Zettlemoyer, and
  Veselin Stoyanov,
\newblock ``Unsupervised cross-lingual representation learning at scale,''
\newblock in {\em ACL}, 2020.

\bibitem{ArT}
Chee~Kheng Chng, Yuliang Liu, Yipeng Sun, Chun~Chet Ng, Canjie Luo, Zihan Ni,
  ChuanMing Fang, Shuaitao Zhang, Junyu Han, Errui Ding, et~al.,
\newblock ``Icdar2019 robust reading challenge on arbitrary-shaped
  text-rrc-art,''
\newblock in {\em ICDAR}, 2019.

\bibitem{ReCTS}
Rui Zhang, Yongsheng Zhou, Qianyi Jiang, Qi~Song, Nan Li, Kai Zhou, Lei Wang,
  Dong Wang, Minghui Liao, Mingkun Yang, et~al.,
\newblock ``Icdar 2019 robust reading challenge on reading chinese text on
  signboard,''
\newblock in {\em ICDAR}, 2019.

\bibitem{wals}
Matthew~S. Dryer and Martin Haspelmath,
\newblock ``Wals online (v2020.3),'' https://wals.info/, 2013,
\newblock Accessed: 2023-09-11.

\end{thebibliography}

\end{document}